\documentclass[conference]{IEEEtran}
\IEEEoverridecommandlockouts
\usepackage{cite}
\usepackage{amsmath,amssymb,amsfonts}
\usepackage{algorithmic}
\usepackage{graphicx}
\usepackage{textcomp}
\usepackage{xcolor}
\usepackage{cuted}
\usepackage{url}
\usepackage{subcaption}
\usepackage{caption}
\usepackage{multirow}
\usepackage{enumitem}
\usepackage[labelfont=bf, labelsep=period]{caption}
\usepackage{amssymb}
\usepackage{pifont}
\newcommand{\cmark}{\textcolor{olive}{\ding{51}}}%
\newcommand{\xmark}{\textcolor{purple}{\ding{55}}}%

\newcommand{\wutterance}{0.35\textwidth}
\newcommand{\wtask}{0.35\textwidth}
\newcommand{\tabhead}[1]{\multicolumn{1}{|c|}{\textbf{#1}}}

\def\BibTeX{{\rm B\kern-.05em{\sc i\kern-.025em b}\kern-.08em
    T\kern-.1667em\lower.7ex\hbox{E}\kern-.125emX}}
\begin{document}

\title{Robot Behavior Personalization from \\ Sparse User Feedback\\
{}
} 

\newcommand{\new}[1]{#1}

\author{\IEEEauthorblockN{Maithili Patel}
\IEEEauthorblockA{\textit{Georgia Institute of technology} \\
\texttt{maithili@gatech.edu}}
\and
\IEEEauthorblockN{Sonia Chernova}
\IEEEauthorblockA{\textit{Georgia Institute of technology} \\
\texttt{chernova@gatech.edu}}
}

\maketitle

\begin{strip}
\begin{minipage}{\textwidth}\centering
\vspace{-35pt}
    \includegraphics[width=0.96\textwidth]{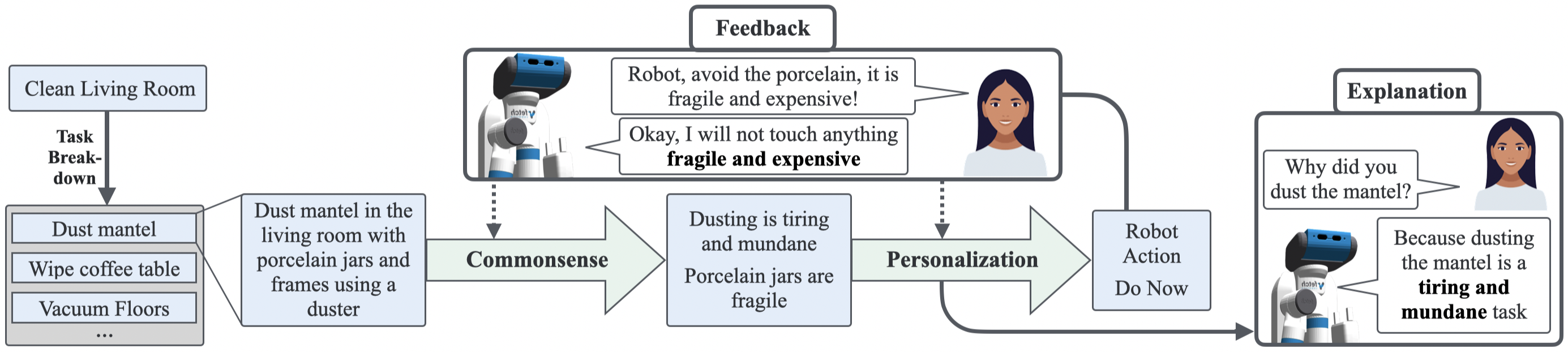}
\setlength{\abovecaptionskip}{0pt}
\setlength{\belowcaptionskip}{-12pt}
\captionof{figure}{\small{Task Adaptation using Abstract Concepts (TAACo) learns user preferences regarding how they want the robot to assist with an open set of household tasks from limited user feedback. This requires commonsense reasoning to extract relevant semantic information regarding a novel task, and personalization based on limited feedback. In addition, TAACo can explain its predictions to the user in an intuitive manner.}}
\label{fig:title}
\end{minipage}
\end{strip}

\begin{abstract}
As service robots become more general-purpose, they will need to adapt to their users' preferences over a large set of \textit{all} possible tasks that they can perform.
This includes preferences regarding which actions the users prefer to delegate to robots as opposed to doing themselves.  
Existing personalization approaches require task-specific data for each user. To handle diversity across all household tasks and users, and nuances in user preferences across tasks, we propose to learn a task adaptation function independently, which can be used in tandem with any universal robot policy to customize robot behavior.
We create Task Adaptation using Abstract Concepts (TAACo) framework. TAACo can learn to predict the user's preferred manner of assistance with any given task, by mediating reasoning through a representation composed of abstract concepts built based on user feedback. TAACo can generalize to an open set of household tasks from small amount of user feedback and explain its inferences \new{through intuitive concepts}.
We evaluate our model on a dataset we collected of 5 people's preferences, and show that TAACo outperforms GPT-4 by 16\% and a rule-based system by 54\%, on prediction accuracy, with 40 samples of user feedback.
\end{abstract}


\section{Introduction}

Service robots have the potential to assist diverse users across warehouses, homes, assisted living facilities, etc., by performing tasks based on user commands~\cite{jiang2022vima, shridhar2023perceiver, lynch2023interactive, liang2024learning}.
However, these methods require unambiguous commands~\cite{jiang2022vima, shridhar2023perceiver}, such as `pick up the cereal box from the pantry and bring it here', or need the user to resolve ambiguities during operation~\cite{lynch2023interactive, liang2024learning}. For complex, high-level tasks, such as `help me with breakfast', unambiguous definition of every detail is impractical, and robots must autonomously adapt to user preferences to assist with under-specified tasks. Such adaptation includes determining `when' a task should be done, `how' it should be done, and `who' should do it.
Prior works have explored questions of anticipating `when' a task should be done~\cite{patel2022proactive, patel2023predicting, mascaro2023hoiabot}, and details of `how' the user prefers a task to be executed~\cite{kapelyukh2022my, jain2023transformers, peng2023diagnosis, yuan2022situ}, but the final question of `who' should do a given task, between humans and robots~\cite{ranz2017capability, malik2019complexity,gjeldum2022collaborative, zhao2023learning}, is relatively under-explored in the context of personal household preferences. This problem involves two main challenges: the diversity in preferences across users, and diversity across the vast space of tasks that a robot might encounter.
Given the growing demand for robotic support in homes and assisted living facilities, we focus on older adults, a demographic that prior studies~\cite{smarr2014domestic} have shown differ in their preferences regarding high-level activities, such as \textit{meal preparation}, that they prefer delegating to a robot.
We examine preferences over more granular tasks, such as \textit{wiping the countertop} and \textit{preparing coffee}, through a case study involving interviews with five older adults, and find low agreement on preferences between users, indicated by an average pairwise Cohen's Kappa of 0.23, underscoring the need for personalization to each user. 

Although users prefer to customize robots themselves~\cite{saunders2015teach},
existing methods on task allocation~\cite{ranz2017capability, malik2019complexity} in human-robot teaming primarily focus on optimizing task efficiency rather than user preferences, and methods that consider user preferences~\cite{gjeldum2022collaborative, zhao2023learning} require exhaustive annotation or prior experience across the space of tasks, which becomes intractable for general robots that can perform a multitude of tasks.
If users were required to set explicit rules for the preferences we gathered, they would require on average 76.5 rules for every 100 scenarios.
Our \textbf{key insight} is that while the space of all actions is diverse and intractable, people's preferences associated with them are anchored in a relatively smaller space of abstract concepts (e.g. fragility of an object, mundaneness of a task, being particular about a task, etc.). Moreover, an understanding of such abstract concepts can be bootstrapped from prior knowledge sources, and utilizing them explicitly in an intermediate representation helps the robot align its reasoning mechanism with the user's, and leverage user-generated explanations to further improve such alignment. Ultimately, this not only helps generalization, but enables such a model to intuitively explain model inferences to the user. 



We contribute TAACo, which leverages prior knowledge to create a semantically-rich low-dimensional intermediate representation to adapt robot behavior to align with user preferences from sparse user feedback. We define robot behavior adaptation with respect to a given task as determining who should perform it, and any communication involved in this allocation. In this regard, TAACo can adapt \textit{any} robot task, generated by \textit{any} universal policy, to the particular user, from small amounts of verbal user feedback on prior tasks.
For instance, if a user prefers the robot to prepare ingredients for an omelet but not cook it, the robot should be able to infer this preference from past feedback, such as the user asking not to cook pasta due to being particular about food, and requesting help with tasks like clearing the kitchen counter after cooking.
For a given task, TAACo can help a robot decide whether it should perform the action, and whether it should do so right now or later, or should it allow the user to execute it, and whether it should remind them about it.
Moreover, TAACo trains only on target user's data, eliminating the need for data sharing, and allowing data to be stored and used locally.

The TAACo framework\footnote{The code for TAACo is available at \url{https://github.com/Maithili/TAACO}}, outlined in Figure~\ref{fig:title}, can learn to predict the preferred robot behavior adaptation for any given task, by reasoning through a space of abstract concepts. Concretely, it offers the following key advancements
\begin{enumerate}
    \item The ability to generalize user preferences to an open set of tasks from a small amount of user feedback
    \item The ability to explain its decisions faithfully through abstract concepts that a user can understand
\end{enumerate}

Evaluations on a custom dataset\footnote{The dataset is available publicly at \url{https://tinyurl.com/taacodataset}} show that TAACo outperforms GPT-4 by 16\% and a rule-based system by 54\%, on accuracy of correctly predicting the preferred manner of assistance based on 40 samples of user feedback.
\begin{table*}[htbp]
  \centering
  \caption{\small{Data collected and parsed from user interviews.}\vspace{-5pt}}
  \label{tab:data}
  \begin{tabular}{|p{0.0005\textwidth}|p{\wutterance}|p{\wtask}|p{0.095\textwidth}|p{0.05\textwidth}|}
    \hline
     &  &  & \tabhead{State} & \tabhead{Task}  \\
     & \tabhead{User Utterance} & \tabhead{Task Description $\boldsymbol{t}$} & \tabhead{Constraint~$S_c$} & \tabhead{Adaptation~$\phi$} \\
    \hline

    \multirow{3}{*}{1} & \multirow{2}{\wutterance}{\textit{``I like doing it myself but I'm usually quite busy, and don't have the patience it requires, so I'd want the robot to do it... Maybe over weekends, when I have more time, it should remind me instead."}} & \multirow{2}{\wtask}{Watering indoor plants with a water jug, Maintaining house plants, \{Water jug, House plants\}, \{Living Room\}} & not weekend & \textit{do\_now} \newline \\ \cline{4-5}
    & & & weekend & \textit{remind} \\ 
    & & & & \\
    \hline

    
    2 & \textit{``Yep, same [as other cooking tasks], it should do it. I don't like cooking."} & Pouring oil in a pan to cook food,	Preparing a Meal,	\{Oil Bottle, Pan, Stove\},	\{Kitchen\} & - & \textit{do\_now} \\
    \hline

    3 & \textit{``No, the robot shouldn't ever operate the stove; it is a serious fire hazard."} & Turning on the stove,	Preparing a Meal,	\{Stove, Stove knobs\},	\{Kitchen\} & - & \textit{no\_action} \\
    \hline

    \multirow{2}{*}{4} & \multirow{2}{\wutterance}{\textit{``Yes, I would love for the robot to do such mundane activities... But it shouldn't cause ruckus when I'm asleep."}} & \multirow{2}{\wtask}{Arranging pots and pans in the kitchen shelves,	Organizing the Kitchen,	\{Pots, Pans, Kitchen Shelves\},	\{Kitchen\}} & user is not asleep & \textit{do\_now} \\ \cline{4-5}
    & & & user is asleep & \textit{do\_later} \\ 
    \hline



    5 & \textit{``Sure, the robot can do it... I trust the robot to follow a recipe when baking."} & mixing cake batter to bake a birthday cake, baking, \{cake batter, mixing bowl, wooden spoon\}, \{kitchen\} & - & do\_now \\
    \hline

    6 & \textit{``No, I'm particular about cooking and such, I'd rather do it myself."} & mixing cake batter to bake a birthday cake, baking, \{cake batter, mixing bowl, wooden spoon\}, \{kitchen\} & - & no\_action \\
    \hline

    \multirow{2}{*}{7} & \multirow{2}{\wutterance}{\textit{``I don't mind the robot folding my laundry... But it is tight around my closet, so if I'm there, I wouldn't like it moving around me so much."}} & \multirow{2}{\wtask}{Folding and putting away clothes in the dresser, Doing Laundry, \{Pants, Shirt, Jackets, Dresses, Dresser\},	\{Closet\}} & user is close by & \textit{do\_later} \\ \cline{4-5}
    & & & user is not close by & \textit{do\_now} \\ 
    \hline
  \end{tabular}
\vspace{-15pt}
\end{table*}
\section{Background and Related Work}

In this section, we discuss prior work on personalizing robot behavior to user preferences and bootstrapping prior knowledge contained in LLMs.

A common approach to personalization of robot tasks is for users to utilize end-user programming methods to directly codify their preferences. Interfaces which allow a user to define `if...then...' style rules based on abstract semantic conditions, such as \textit{food is being prepared}~\cite{saunders2015teach}, or IoT sensor triggers~\cite{leonardi2019trigger}, have enabled personalized scheduling of high level behaviors. Automatic parsing of user rules into Temporal Logic has been used to administer customized cognitive therapy sessions~\cite{kubota2020jessie}. However, in giving the users total control over robot actions, these methods require the user to be accurate and exhaustive in defining their expectations. For complex behaviors involving detailed preferences, user-specification of tasks have been shown to fail due to unforeseen repercussions~\cite{zhuang2020consequences, skalse22defining}. 


More generally, personalization in human robot interaction involves adapting to user's task preferences on `how' and `when' a task should be executed, as well as `who' should execute a given task.
Towards addressing the `how' question, prior works have learned to adapt to the user's preferred manner of executing various tasks, such as learning preferred trajectories, object locations, or action ordering, based on demonstrations~\cite{peng2023diagnosis,kapelyukh2022my, jain2023transformers}, interactive feedback~\cite{cui2023no, sharma2022correcting, yuan2022situ}, facial expressions~\cite{cui2020empathic}, or scene context~\cite{ramachandruni2023consor, lindner2021learning}.
Prior works have also explored the `when' question to enable proactivity through behavior prediction. In collaborative settings~\cite{mascaro2023hoiabot, munzer2017preference, zhao2022coordination, nemlekar2023transfer}, such as robot handovers, short-term predictions have been used to improve task efficiency and fluency, and in household assistance~\cite{patel2022proactive, patel2023predicting} long-term predictions have enabled timely robot assistance without direction.

The final question is that of accounting for user preferences while determining `who' a task should be allocated to; this topic remains relatively under-explored in prior work in the context of household robots.  The field of task allocation has extensively addressed assignment of tasks in multi-robot and human-robot teams~\cite{ranz2017capability, malik2019complexity, gjeldum2022collaborative}.  However, existing works primarily weigh the utilities and costs of various allocations, aiming to optimize the efficiency of achieving task goals.
While this is useful in factory settings, in domestic settings, the goal shifts from task efficiency to promoting user comfort and satisfaction. In such settings, the robot must understand allocation preferences of which tasks the users \textit{prefer} doing themselves, as opposed to delegating to a robot.
Closest to our work are methods that seek to recover hidden human preferences, by learning a reward function~\cite{zhao2023learning, hadfield2016cooperative}. But these works are limited to pre-specified tasks and require past collaboration experience with the particular user on those tasks.
Transferring user preferences across tasks has been investigated towards optimizing handovers~\cite{nemlekar2023transfer}, and expanding user capabilities~\cite{du2020ave}.
We aim to generalize allocation preferences of the user across an open set of tasks.

Finally, abstract semantic knowledge from foundational models has been leveraged in various robotic applications, but most related to our work is task-related robotic assistance for humans. Foundational models have enabled direct user interaction with robots towards efficiently teaching tasks to the robot~\cite{liang2024learning}, natural instruction following~\cite{quartey2024verifiably, song2023llm}, creating queryable scene representations~\cite{chen2023open, huang2023visual}, and explaining robot failures~\cite{sobrin2024explaining}. Knowledge about human norms encoded in Large Language Models (LLMs) has been leveraged to personalize robot behavior~\cite{wu2023tidybot} and to model humans~\cite{zhang2023large}.
These works use the LLM as a reasoning mechanism to directly perform the target task~\cite{zhang2023large, sobrin2024explaining} or create a structured framework to solve the task and employ the LLM to solve parts of it~\cite{wu2023tidybot, liang2024learning}. 
Prior works have also used LLMs to embed input information into semantically rich latent spaces~\cite{shridhar2023perceiver, brohan2022rt1}. In addition to using latent embeddings, we use an LLM to create an intermediate representation explicitly in the form of abstract concepts.

\section{Problem Formulation}
\vspace{-5pt}

\new{The primary aim of this work is to learn a model $\Phi$ to predict the desired task adaptation $\phi$ to regulate the execution of $\boldsymbol{t}$ to match user preferences, and produce an explanation $\mathcal{E}_{robot}$ for its prediction. We assume that the high-level behavior of the robot is governed by a plan or policy $\pi$, which suggests task $\boldsymbol{t}$ that the robot should perform, in accordance with the world state $ \boldsymbol{s} \in S$, and robot's goal. As the robot performs various tasks $\boldsymbol{t}$ in the household, produced by $\pi$ either proactively or reaction to a command, the user can provide feedback $u$, about their preferred adaptations. Over time, using $u$, the robot must learn a function $\Phi$ to predict the preferred task adaptation $\phi$ for novel tasks that it encounters.}


We represent task $\boldsymbol{t}$, through a tuple of the components describing it, $\boldsymbol{t} = (c^a, c^h, \{c^o\}, \{c^l\})$. $c^a$ is an action description, e.g. \textit{dusting the mantel}. $c^h$ is the corresponding high-level activity, e.g. \textit{cleaning the living room}. $\{c^o\}$ is a set of objects involved in the action, e.g. \textit{\{mantel, duster, porcelain jar, photo frame\}}. $\{c^l\}$ is a set of locations involved in executing the action, e.g. \textit{\{living room\}}. \new{We allow each component to be any natural language phrase and so the task is not confined to a closed set, and can be generated by open-set language-based planners, as well as classical or RL-based planners for which language labels are available.}
The state $ \boldsymbol{s} \in S $ is comprised of a set of binary variables $ \boldsymbol{s}  = \{ s_i \} $, and can include predicates such as `user is asleep', `guests are present', `weekend' etc.

The aim of this work is to utilize prior user feedback $u$ to adapt robot behavior. We use a task adaptation $\Phi(\boldsymbol{t}, \boldsymbol{s}) \rightarrow \phi$ to represent adaptation of robot behavior over a task $\boldsymbol{t}$.
The task adaptation $\phi$, can be one of acting on task $\boldsymbol{t}$ immediately (do\_now), acting on $\boldsymbol{t}$ later (do\_later), reminding or requesting the user to do $\boldsymbol{t}$ (remind), and not doing anything about $\boldsymbol{t}$ (no\_action). A function $\Phi$ which predicts a task adaptation is used to personalize robot behavior in one of two ways: as a filter downstream of $\pi$ to determine whether or not to execute the action selected by $\pi$, or as an external constraint to $\pi$, allowing it to optimize for user preferences, similar to accounting for user schedules~\cite{booth2016mixed} or motion constraints~\cite{dantam2016incremental}.

To personalize to a given user, the robot has access to user feedback $u$ over prior tasks (examples shown in in Table~\ref{tab:data}). User feedback samples $(\phi, \boldsymbol{t}, S_c, \mathcal{E}_{user})$ include the preferred task adapter $\phi$ for a task $\boldsymbol{t}$, and can optionally include a state constraint $S_c$ and user-generated explanation $\mathcal{E}_{user}$. The state constraint $S_c$ identifies a subset of the state space where the preference applies. The user-generated explanation, (examples shown in Table~\ref{tab:outputs}), expresses the user's reasoning behind the given preference $\mathcal{E}_{user} = \{(c^x, \theta^x)\}$, and includes a set of abstract concepts $\theta^x$ associated with components $c^x$ of the task, for any type of component $x\in \{ a,h,o,l \}$. Similar to the user generated explanation, the robot should also be able to offer an explanation $\mathcal{E}_{robot} = \Psi(\Phi, \boldsymbol{t}, \boldsymbol{s})$ for its prediction.

\begin{figure}[h]
    \centering
    \vspace{-10pt}
    \includegraphics[width=0.48\textwidth]{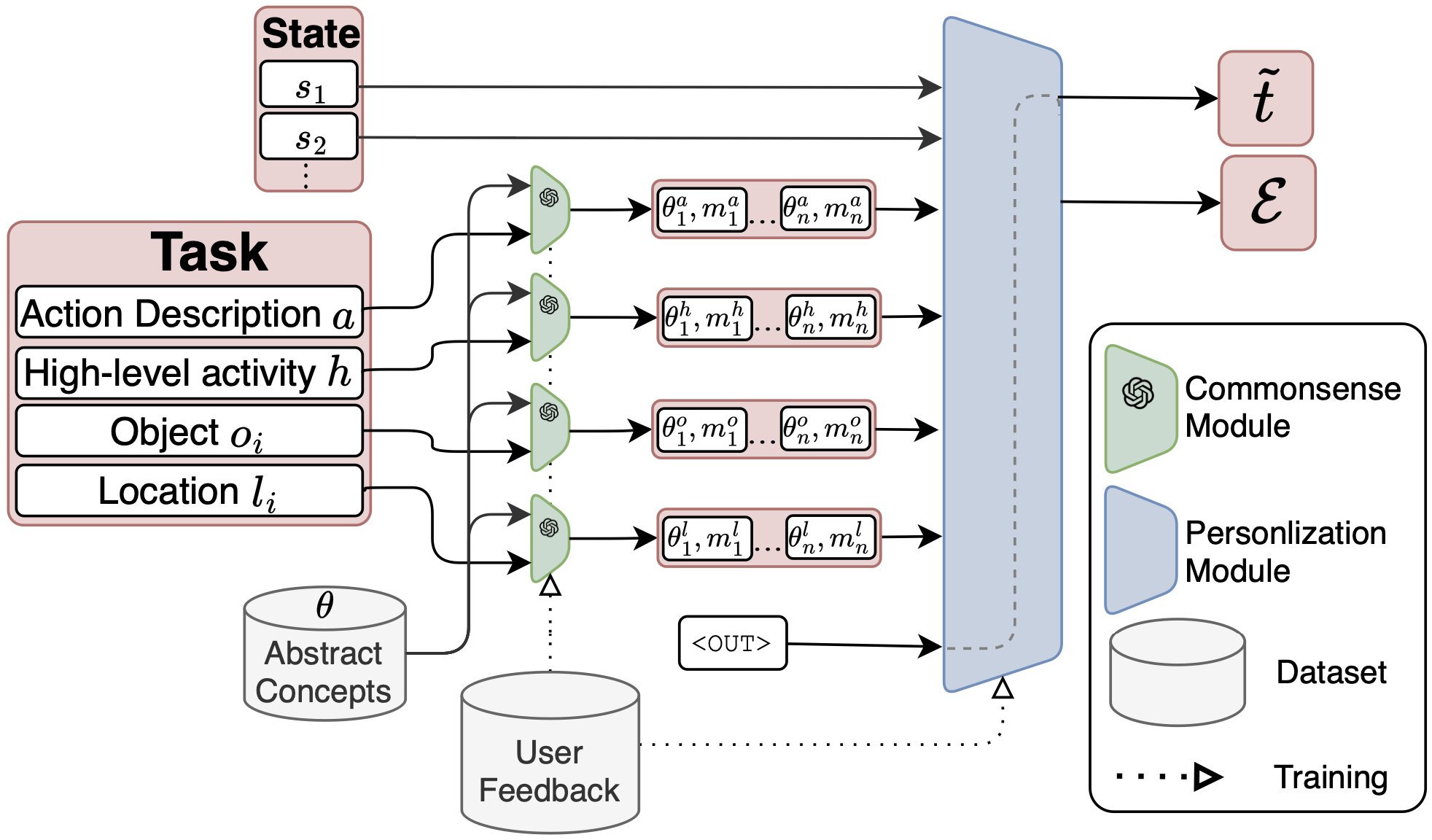}
    \setlength{\abovecaptionskip}{-13pt}
    \setlength{\belowcaptionskip}{-13pt}
    \caption{\small\new{TAACo generates a concept-based representation for a given task through a Commonsense Module, and uses it to predict the preferred action adaptation through a Personalization Module.}}
    \label{fig:system}
\end{figure}

\section{Method}


The Task Adaptation using Abstract Concepts framework (TAACo), addresses the two main challenges in learning $\Phi$, portrayed in Figure~\ref{fig:title}: 1) generalization to an open-set of tasks by extracting relevant semantic information from any novel household task, and 2) personalized prediction of preferred action adaptation based on limited feedback. As outlined in Figure~\ref{fig:system}, TAACo is composed of a Commonsense module and a Personalization module to address the two challenges. Based on the idea that people's preferences are grounded in abstract semantic concepts, such as \textit{fragility} of objects or \textit{mundaneness} of tasks, we use such semantic concepts to create an intermediate representation between the two modules. This representation helps explain model decisions in an intuitive manner, and leverage user-generated explanations to align the personalization model with the user. Unlike prior works \cite{das2023state2explanation, zabounidis2023concept}, we do not restrict the set of concepts to a predefined closed set, instead allowing the user to define new concepts as needed through natural language. 

The aim of the \textbf{Commonsense Module} is to extract relevant semantic information about the task $\boldsymbol{t}$ into an intermediate representation $\tilde{t} = \{ (x, \theta^x, m) \}$. 
To create $\tilde{t}$ we distill semantic information from each task component $c^x$, which can be of one of four component types $x \in \{a,h,o,l\}$ (action, high level activity, object, location), by predicting a score $m$ of how well it matches each of a set of relevant abstract concepts $\{\theta^x\}$. 
For every concept $\theta^x$, component $c^x$ pair, we prompt GPT-4 to predict a score on a scale of 1-10, and linearly rescale it to [0,1], to obtain $m$.
Each component-concept pair prompt is independent, without any context of other task components, concepts or prior feedback, allowing responses to
be cached and reused across personas and tasks.
We create an initial set of concepts for each component type, emulating a factory configuration, and adapt to each user by adding new concepts obtained from user feedback to these sets.

The \textbf{Personalization Module} predicts the preferred adaptation for each task, given the task and world state. The task input is represented in the form of abstract concepts $\tilde{t} = \{ (x, \theta^x, m) \}$, and the world state as a set of binary variables $S = \{s_i\}$.
We first embed each input, including each tuple element of the task representation and state variables, independently into a set of three latent vectors: a type embedding~$h_x$, a concept embedding~$h_\theta$, and a score embedding~$h_m$. The three vectors are concatenated to form inputs to a transformer encoder with a classification head to predict the final task adaptation.

For the task representation, we embed each element of the tuple $(x, \theta^x, m)$ to create the three latent vectors. We create $h_x$ by learning an embedding for each component type $x \in \{a,h,o,l\}$. We create $h_\theta = \rho_c(\gamma_{LLM}(\theta))$ using an off-the-shelf language embedder $\gamma_{LLM}$, and projecting the resulting embedding into latent space using a learned concept projection function $\rho_c$. We create $h_m = \rho_m(m)$ using a feedforward magnitude projection function $\rho_m$ to project the match scores $m$. To embed the task representation, we use a single learned $h_x$, and a learned concept embedding $h_\theta$ for each binary state variable $ s_i \in S $. To create $h_m$ reuse $\rho_m$ to project the binary state value $h_m = \rho_m(m), m \in \{ 0,1 \}$.
We pass the final latent embedding for each input component $h_x|h_\theta|h_m$, along with an output token, \texttt{<OUT>}, through a transformer encoder. We use the output feature at \texttt{<OUT>} to classify the desired task adaptation through a classification head. We train the Personalization Module in a supervised manner using preference data obtained from the user.


We \textbf{generate model explanations} using the human-interpretable intermediate representation, and \textbf{align the model's reasoning with the user's} by explicitly training the model's explanations.
Such training not only improves the model's ability to explain, but also improves its ability to generalize to new actions, by learning the correct underlying structure of the data and avoiding spurious correlations. Results in Section~\ref{sect:ablations} empirically verify this claim.

We define the space of all explanations to be the set of all inputs to the Personalization Module, namely all component-concept pairs that form the task representation and all state variables $\mathcal{E} \in \{(c^x, \theta^x)\} \cup \{s\}$. Hence, extracting an explanation is equivalent to finding the most important component in the input to the Personalization Model. We use attention weights $w_a$ between the output token \texttt{<OUT>} and each input to compute the probability of that input component to be the explanation as $1-e^{-w_a}$. The input component with the highest probability is used as the explanation.
We add an auxiliary loss to motivate the explanation probabilities to align with user feedback, through a binary cross entropy loss evaluated based on whether each input component is a part of explanations $\mathcal{E}_{user}$ or constraints $s \in S_c$ provided by the user for that task.
We use attention weights $w_a$ from the final layer of the transformer encoder to allow context from other inputs to influence the attention weights through previous layers.

\begin{table*}[htbp]
  \centering
  \caption{\small{Examples of predictions and explanations generated by TAACo and baselines.}\vspace{-5pt}}
  \label{tab:outputs}
  \begin{tabular}{|p{0.001\textwidth}|p{0.15\textwidth}|p{0.07\textwidth}|p{0.2\textwidth}|p{0.16\textwidth}|p{0.17\textwidth}|p{0.05\textwidth}|}
    \hline
    & \tabhead{Task $\boldsymbol{t}$} & \tabhead{State $\boldsymbol{s}$} & \tabhead{Ground Truth} & \tabhead{TAACo} & \tabhead{GPT} & \tabhead{RuleBased} \\
    \hline
    1 & 
     putting fruits in the blender to make a smoothie, Making a smoothie, \{blender, kitchen counter, apple, banana, strawberry\}, \{kitchen\} & user is in a rush \& weekend & 
     $\phi$: no\_action \newline $\mathcal{E}$: putting fruits in the blender to make a smoothie is an action that a user might prefer doing themselves if they enjoy making food. & 
     $\phi$: no\_action~\cmark \newline $\mathcal{E}$: putting fruits in the blender to make a smoothie is an action that a user might prefer doing themselves if they enjoy making food.~\cmark & 
     $\phi$: do\_now~\xmark \newline $\mathcal{E}$:  Making a smoothie is a/an activity which is falls under food preparation tasks.~\xmark & 
     $\phi$: do\_now~\xmark \newline $\mathcal{E}$:  -~\xmark\\
    \hline      
    2 & 
    drilling holes in the wall to put up a coat hook, home decoration, \{electric drill, hammer, screws\}, \{living room\} & adverse weather conditions \& guests are present \& weekend & 
    $\phi$: no\_action \newline $\mathcal{E}$:  drilling holes in the wall to put up a coat hook is an action that can cause major damage or harm if done imprecisely, and electric drill is an object that can easily hurt someone without intending to & 
    $\phi$: no\_action~\cmark \newline $\mathcal{E}$:  drilling holes in the wall to put up a coat hook is an action that can cause major damage or harm if done imprecisely~\cmark & 
    $\phi$: no\_action~\cmark \newline $\mathcal{E}$:  Drilling holes in the wall to put up a coat hook is a/an action which can cause major damage or harm if done imprecisely~\cmark & 
    $\phi$: no\_action~\cmark \newline $\mathcal{E}$:  Object electric drill is involved~\cmark \\
    \hline      
    3 & 
     arranging pots and pans in the kitchen shelves, organizing the kitchen, \{pots, pans, kitchen shelves\}, \{kitchen\} & user is asleep \& weekend & 
     $\phi$: do\_later \newline $\mathcal{E}$:  Arranging pots and pans in the kitchen shelves is an action that makes a lot of noise, and user is asleep & 
     $\phi$: no\_action~\xmark \newline $\mathcal{E}$: Pot is an object which involves an open flame~\xmark & 
     $\phi$: do\_now~\xmark \newline $\mathcal{E}$: Arranging pots and pans in the kitchen shelves is a/an action which is is very tiring~\xmark &  
     $\phi$: do\_now~\xmark \newline $\mathcal{E}$:  -~\xmark\\
    \hline      
    4 & 
     watering house plants, maintaining house plants, \{watering can, house plants\}, \{living room\} & user is in a rush & 
     $\phi$: do\_now \newline $\mathcal{E}$: watering house plants is a task that a user might want to be carried out in a particular manner if they are particular about their house plants, and the user is in a rush & 
     $\phi$: do\_now~\cmark \newline $\mathcal{E}$: user is in a rush~\cmark &
     $\phi$: do\_now~\cmark \newline $\mathcal{E}$: Watering house plants is an activity under maintaining house plants, which is a mundane chore that robots can assist with effectively~\xmark &
     $\phi$: do\_now~\cmark \newline $\mathcal{E}$:  -~\xmark \\
     
    \hline
  \end{tabular}
\vspace{-15pt}
\end{table*}

\section{Evaluation}

We conduct quantitative evaluation of TAACo on custom data obtained from real users. In this section, we describe evaluation data, baselines and metrics, and model training.

\vspace{-5pt}
\subsection{Case Study}
\vspace{-2pt}

To evaluate TAACo on real user data, we conducted an IRB approved case study about older adults' preferences across various household tasks through in-depth interviews with five people\footnote{Gender and age ranges: M 80-85, M 70-75, F 70-75, F 65-70, F 75-80.}. The resulting documentation contains preferences of each person across 74-85 household tasks each, adding up to 173 distinct household tasks.
All tasks have a unique action description, are associated with one of 60 high level activities, and utilize 265 objects and 14 locations in the household.
The interview was based on scenarios, such as `You asked the robot to help you prepare a meal... the robot is about to pour oil in a pan on the stove.' In context of each scenario, we asked four questions: 1) `In the given scenario, how would you prefer Stretch to respond?', 2) `Why? Please provide a reason why you picked the above.', 3) `Would you respond differently in certain conditions and how?', and 4) `In what other scenarios would you want Stretch to behave in this manner?'. We parse their responses into a set of datapoints $\{(\phi, \mathcal{E}_{user},  S_c)\}$, as shown in Table~\ref{tab:data}~and~\ref{tab:outputs}.
To ensure that tasks are relevant to the participants' daily lives and are not confined to a closed set, we allow the participants to skip or add tasks. We extract concepts from freeform text, to avoid restricting participants to a fixed concept vocabulary.
Ultimately, for each persona, this process results in a set of tasks, along with the preferred robot behavior type, and optionally an explanation and state constraints. We find that the resulting preferences vary significantly across personas (low average pairwise Cohen's Kappa of 0.23), and across tasks (76.5 rules for every 100 scenarios).


\subsection{Baselines}
\vspace{-3pt}

We benchmark TAACo against two baselines: end-to-end GPT-4~\cite{achiam2023gpt} and a rule-based system~\cite{saunders2015teach}, and also compare against an oracle commonsense version with privileged access to concept matches. 
The GPT-4 baseline, uses a few-shot prompting approach. The prompt includes a full history of interactions including user responses and explanations from the training set, and asks for a response to a novel evaluation task.
Inspired by end-user programming methods~\cite{saunders2015teach}, the rule-based baseline creates explicit rules to predict desired behavior conditioned upon a component of the task and, optionally, a state variable, from explanations present in the training set. During inference, we retrieve applicable rules, and use majority vote to predict the response. If no rules are applicable, we select a default response. Finally, the oracle commonsense version of our model corrects match scores in the intermediate task representation using the ground truth explanations provided by the user, wherever available. This tests our personalization model's performance, assuming it has access to a perfect user-aligned scores, particularly those that are deemed important by the user.

\vspace{-5pt}
\subsection{Metrics}
\vspace{-2pt}

We evaluate prediction and explanation accuracy of TAACo on our dataset. We measure \textbf{prediction accuracy} as the fraction of tasks where predicted label matches the ground truth label, and \textbf{explanation accuracy} as the fraction of tasks for which the top explanation offered by a model is a part of the ground truth set. For each model, we measure explanation accuracy only for tasks where ground truth explanations are available, and the model's prediction is correct, to avoid penalizing models for explaining their wrong answers incorrectly. 
Our explanation accuracy metric evaluates the top explanation alone, and not whether all ground truth explanations are generated, measuring whether the robot's response aligns with the user's expectation if it uses the top explanation.

\begin{figure*}[t]
    \centering
    \begin{subfigure}[t]{0.37\textwidth}
    \includegraphics[width=\textwidth]{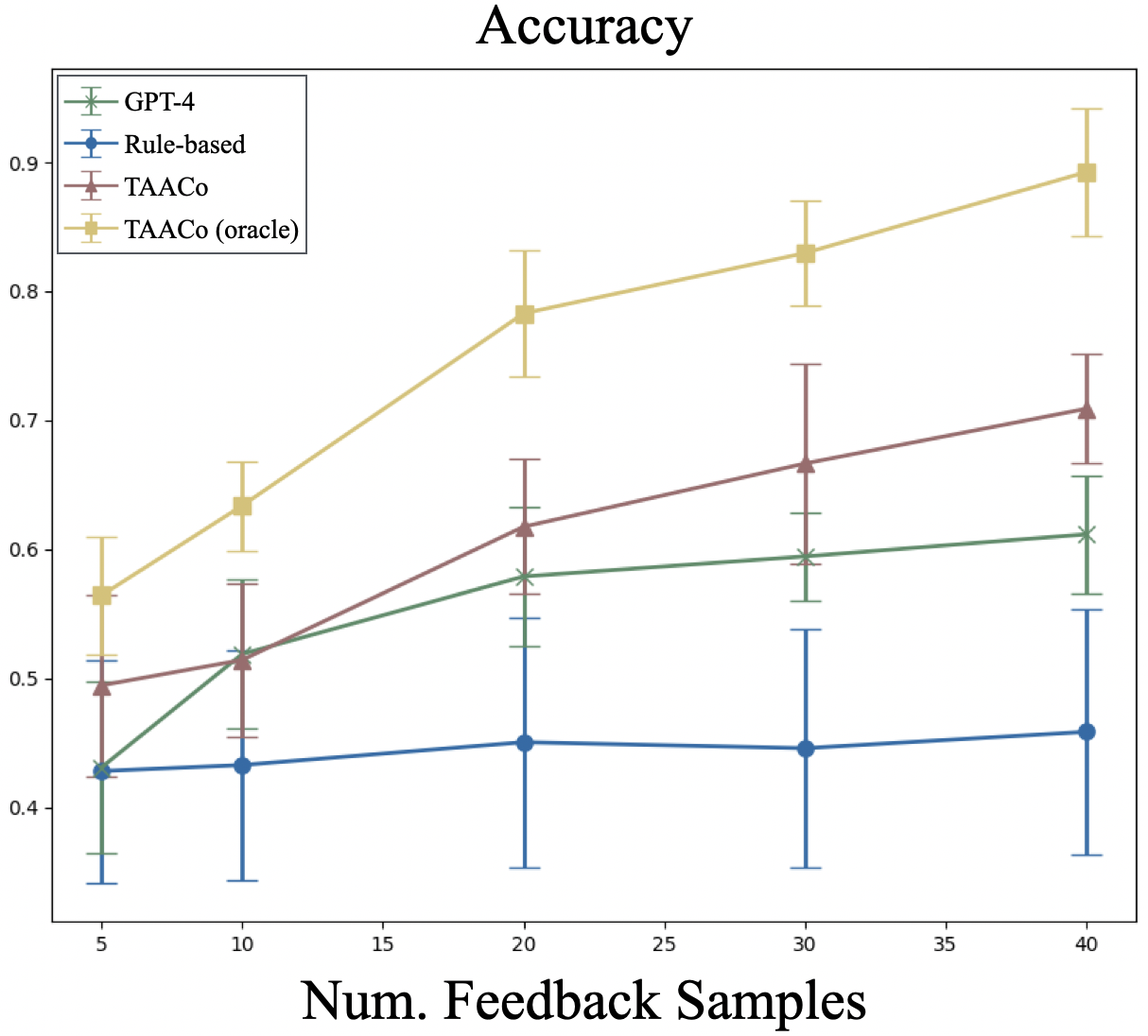}
    \setlength{\abovecaptionskip}{-14pt}
    \setlength{\belowcaptionskip}{-3pt}
    \caption{
    }
    \label{fig:pred_accuracy}
    \end{subfigure}
    \hfill
    \begin{subfigure}[t]{0.61\textwidth}
    \includegraphics[width=\textwidth]{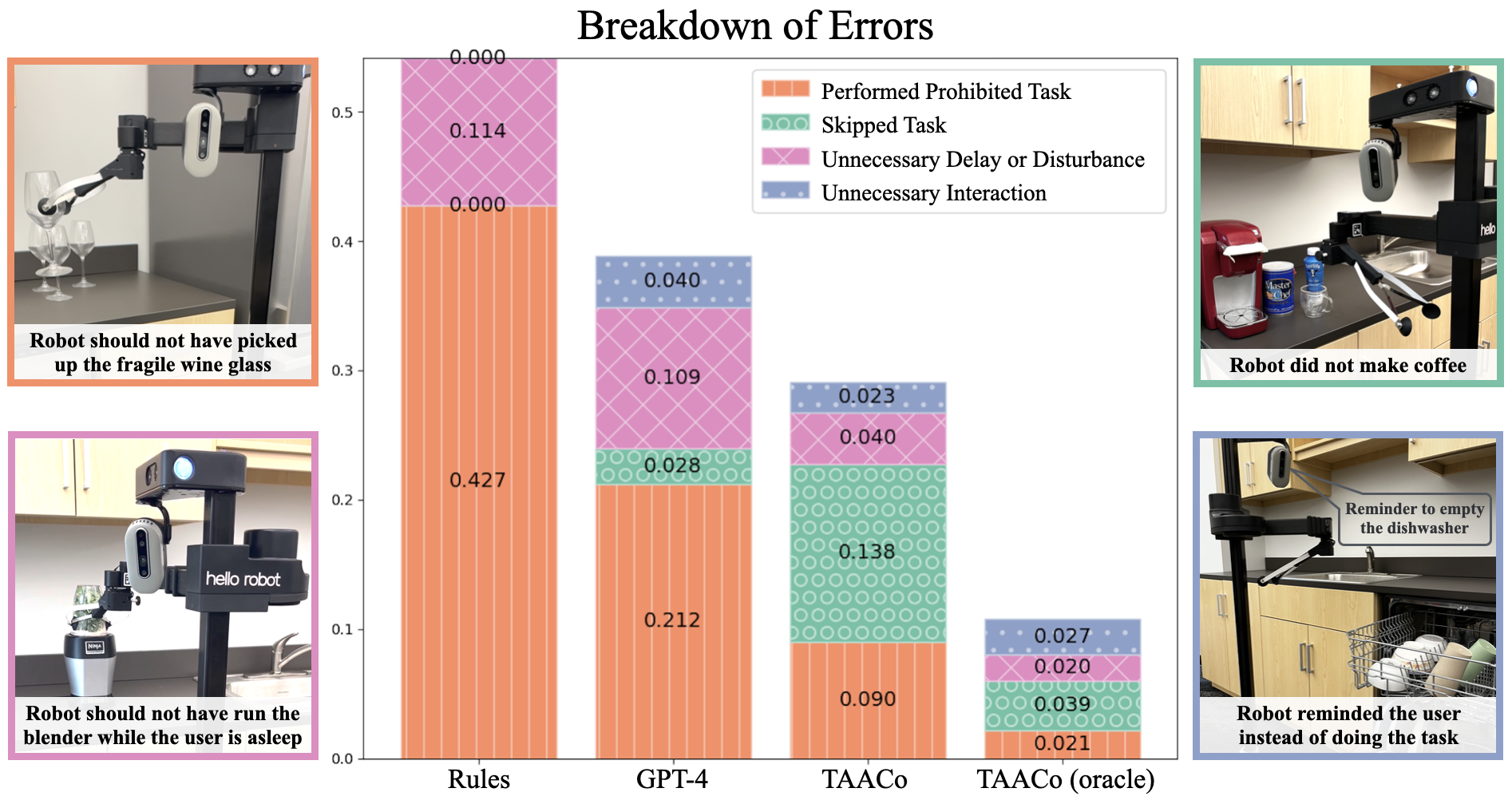}
    \setlength{\abovecaptionskip}{-14pt}
    \setlength{\belowcaptionskip}{-3pt}
    \caption{
    }
    \label{fig:error_analysis}
    \end{subfigure}
    \setlength{\abovecaptionskip}{0pt}
    \setlength{\belowcaptionskip}{-15pt}
    \caption{\small{Comparison of our model against GPT and rule-based baselines, as well as against using an oracle version of commonsense, on (a) overall prediction accuracy over varying amounts of user feedback, and (b) a breakdown of the errors, with 40 feedback samples, with the each bars representing the total error rate and each color representing an error type, along with an example in a box of the same color}}
\end{figure*}

\vspace{-5pt}
\subsection{Model Implementation and Training}
\vspace{-2pt}

All models, including TAACo as well as the baselines, are person-specific, and are trained and tested individually on each person's data using k-fold cross-validation. We evaluate performance across varying user feedback amounts by randomly selecting subsets of the training data. Starting with a base concept vocabulary, we expand it per person based on feedback, resulting in 23-27 concepts each. The state representation consists of 8 binary variables, combining all variables that participants used to define their preferences.

The Personalization Module uses 32-dimensional embeddings for type, concept, and score, creating 96-dimensional inputs for a 2-layer transformer encoder. We use Roberta sentence embeddings as the language encoder $\gamma_{LLM}$. The model, with 909K parameters, is trained for 1700 epochs with the Adam optimizer and a learning rate of $10^{-4}$. We employ a linear combination of cross-entropy loss for predictions and auxiliary explanation loss for attention weights, with the latter given a weight of 20 against unit weight for the former.

\section{Results}

In this section, we compare the prediction and explanation performance of TAACo against baselines and an oracle version of our model, and perform an analysis of the kinds of errors each model makes. We study how performance improves as more user feedback becomes available, by varying the number of user preference samples used in training, with each sample consisting of data pertaining to one task. Finally, we empirically show the importance of using concepts and training based on user-generated explanations. 

\vspace{-5pt}
\subsection{Prediction Accuracy}
\vspace{-2pt}

Figure~\ref{fig:pred_accuracy} shows a comparison of prediction accuracy across 10 to 40 labels obtained as user feedback. When little user data is available, TAACo's performance is similar to GPT-4; TAACo outperforms both baselines as more samples become available. Owing to the vast commonsense knowledge, GPT-4 has a strong commonsense prior which is useful in guessing common preferences, but as some feedback becomes available, TAACo can learn nuances of the user's preferences better, despite GPT-4 having access to the exactly the same information.

Generalization to novel tasks is difficult because each new task has little direct overlap with those seen previously. This is evidenced in the rule-based baseline, which fails to improve with more examples because very few rules created from past examples are applicable to novel situations. Even with 40 training samples, it can only match learned rules for 17\% responses, while the default `do\_now' happens to be correct for the remaining 29\% correct tasks (e.g. row 4 in Table~\ref{tab:outputs}).

TAACo achieves an accuracy of 0.71, while the oracle commonsense version achieves 0.89. This gap in performance results from misalignment between concept scores generated by the GPT-based commonsense module and that provided by the user. Despite the vast commonsense knowledge, GPT-4 can produce unusual or erroneous predictions, such as that the object `pot' involves an open flame (row 3 in Table~\ref{tab:outputs}). Also, there could be subjective differences between people's notions, such as whether being particular about food is applicable to mixing cake batter (rows 5,6 in Table~\ref{tab:data}), causing the universal commonsense model to fail for some persona.

Even with the oracle commonsense, the model can fail due to lack of data or limited expressiveness of our representation. TAACo sometimes fails to learn a complex dependency on multiple variables which occur less frequently, such as task involving moving around, when the location is a tight space, and the user is present (row 7 in Table~\ref{tab:data}). In rare cases, the preference may have never been seen in the train set. For instance, a person didn't want the robot to touch their pet, but neither of the two pet grooming tasks were seen in the training set. Other times, failures can occur when user preferences depend on details which can not expressed in our task representation, such as the difference in the role of the object stove in pouring oil in a pan on the stove, compared to turning on the stove (row 2,3 in Table~\ref{tab:data}).

\vspace{-5pt}
\subsection{Error Analysis}
\vspace{-2pt}

While prediction accuracy reflects overall model performance, different kinds of errors can carry different costs.
To further investigate mistakes made by each model, we categorize errors, based on the their practical impact, as follows.
\begin{enumerate}[leftmargin=*]
    \item \textit{Performed Prohibited Task} when the robot classified a task as \textit{do\_now} or \textit{do\_later}, but was supposed to be \textit{no\_action} or \textit{remind}, implying that the robot would perform a task which the user meant to prevent it from executing.
    \item \textit{Skipped Task} when the robot misclassifies anything as \textit{no\_action}, risking the task remaining unfinished as the user might be expecting the robot to do it or remind them of it.
    \item \textit{Unnecessary Delay or Disturbance} when the robot misclassifies one of \textit{do\_now} and \textit{do\_later} as the other, disturbing the user or delaying the task unnecessarily.
    \item \textit{Unnecessary Interaction} when the robot wrongly predicts \textit{remind}. While a system can be designed to allow the user to correct the robot, it involves interaction with the user.
\end{enumerate}

Note that the four categories are mutually exclusive and exhaustively cover all errors. Figure~\ref{fig:error_analysis} shows the distribution of errors made by each model, when trained on 40 samples, along with an example of each error type. The total height of each bar is the total error rate, and the colors represent the rate of each error type. Notably, most errors for the both baselines are of the least desirable type: \textit{performed prohibited task}. 
The rule-based system's default `do\_now' action, helps prediction performance, due to the prevalence of `do\_now' actions, but causes majority of its errors to fall under this category.
In contrast, TAACo only makes the \textit{performed prohibited task} error 9\% of the time, while a majority of its errors fall under \textit{skipped task}. Overall, this implies that a robot using our model will be more conservative and err towards not acting when uncertain, rather than executing tasks which a user might not want it to do. Our method also shows a lower rate of unnecessary delay or disturbance and unnecessary interaction errors, compared to GPT-4, reducing smaller inconveniences such as interrupting the user by asking questions and making a ruckus, and adding to its conservative nature.

\vspace{-5pt}
\subsection{Explanation Quality}
\vspace{-2pt}

TAACo's explanations constitute a task component and associated concept, matching the ground truth explanation format, but our baselines' explanations do not. The rule-based baseline can use the applied rule to offer the task component as explanation, as seen in row 2 in Table~\ref{tab:outputs}, but has no understanding of concepts, so we use a relaxed metric which deems an explanation to be correct if the right component is identified. For the GPT-4 baseline, despite prompting the system to generate an explanation in the given format, we were unable to fully constrain its responses (e.g. row 1 in Table~\ref{tab:outputs}). So we employed GPT-4 to evaluate its own responses against ground truth through pairwise similarity queries. 

\begin{figure}[h]
    \centering
    \vspace{-7pt}
    \includegraphics[width=0.39\textwidth]{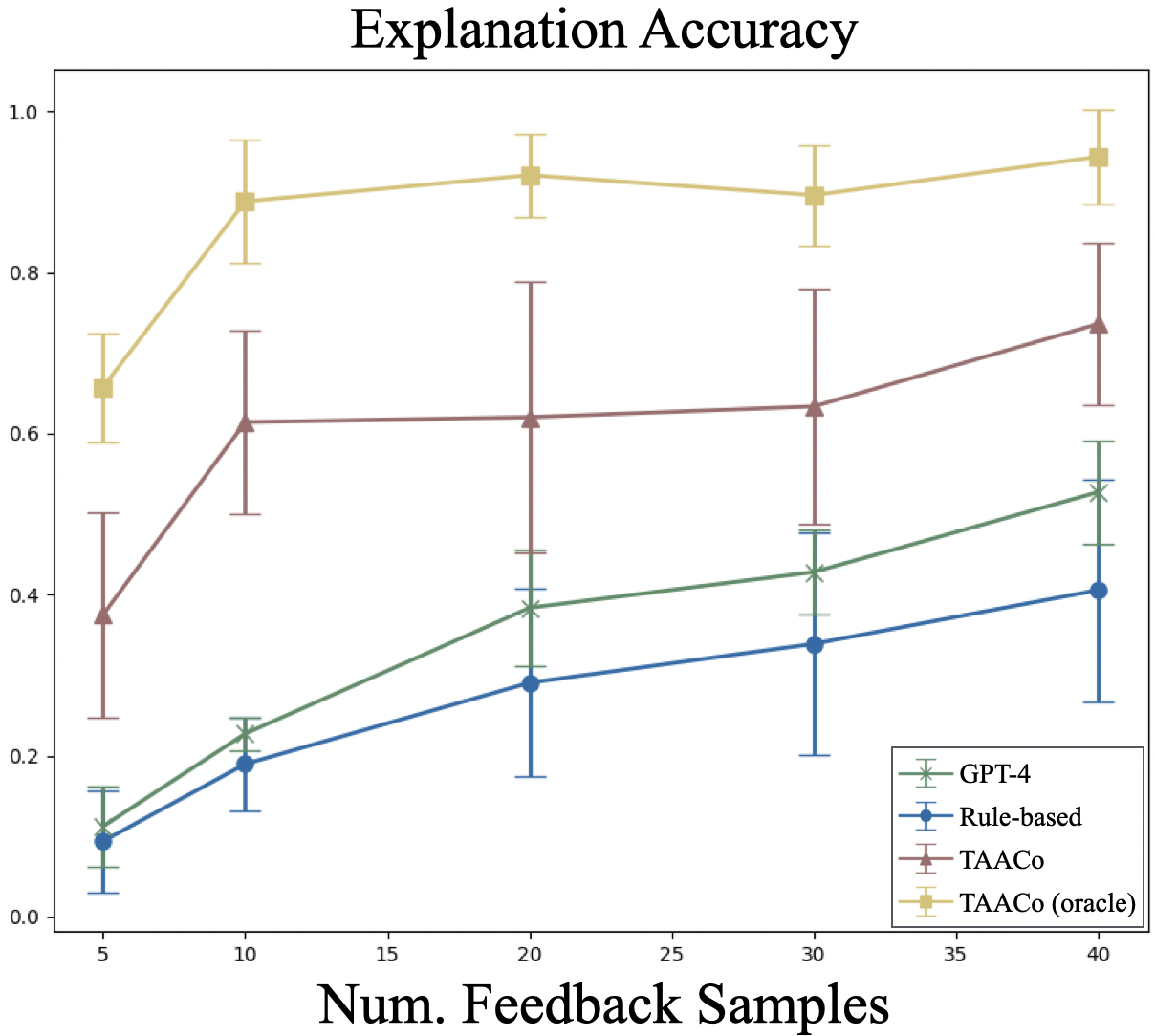}
    \setlength{\abovecaptionskip}{0pt}
    \setlength{\belowcaptionskip}{-8pt}
    \caption{\small{Comparison of Explanation Accuracy against baselines.}}
    \label{fig:expl_accuracy}
\end{figure}

Figure~\ref{fig:expl_accuracy} shows a comparison of explanation accuracy of our model against both baselines and the oracle version, across varying levels of feedback. Our model outperforms both baselines, reaching an accuracy of 0.74 compared to 0.53 and 0.41 by GPT-4 and Rule-based conditions, respectively. The rule based baseline's inability to generalize rules from past feedback causes its explanation performance to be the lowest, despite a significantly relaxed evaluation metric.

\vspace{-5pt}
\subsection{Ablations}
\label{sect:ablations}
\vspace{-2pt}

\begin{figure}[h]
    \centering
    \vspace{-5pt}
    \includegraphics[width=0.49\textwidth]{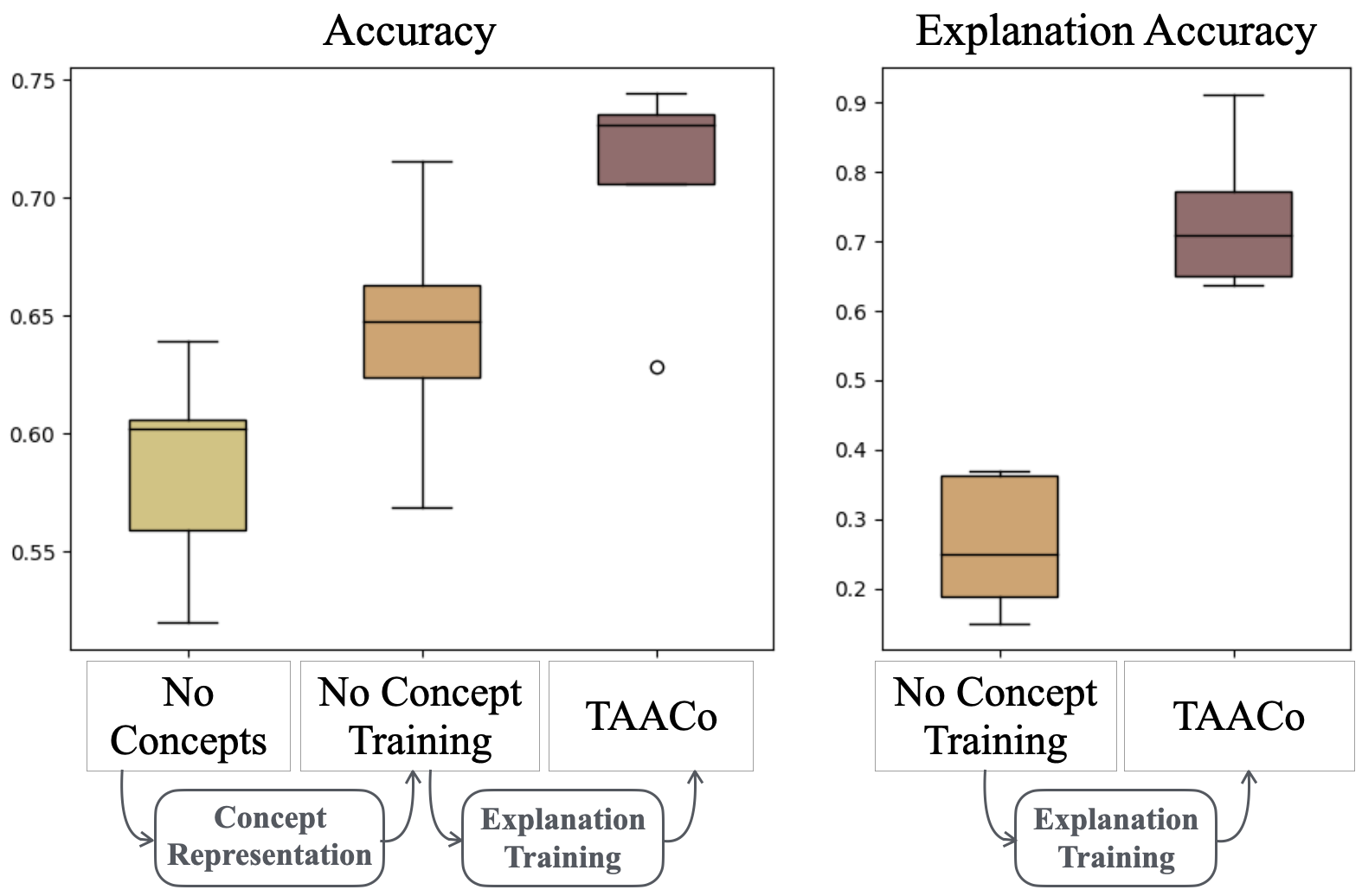}
    \setlength{\abovecaptionskip}{-10pt}
    \setlength{\belowcaptionskip}{-10pt}
    \caption{\small{Ablation results of our model, removing explanation training, and removing the use of concepts entirely.}}
    \label{fig:ablation}
\end{figure}

The core advancement of our model is the explicit use of abstract concepts, both for mediating reasoning and for explicit supervision from user feedback. Figure~\ref{fig:ablation} summarizes performance benefits of both these elements on prediction and explanation accuracy, when trained on 40 feedback samples. 

Our first hypothesis was that mediating reasoning through abstract concepts aids generalization by bootstrapping more relevant prior knowledge than directly using language embeddings. To test this, we compare against a `No Concepts' version of our model which directly uses language embeddings of each component, instead of explicit concepts, as inputs to the Personalization Module, replacing the concept embedding and match score embedding. The no-concept model achieves only 58.5\% prediction accuracy, compared to the 70.9\% achieved by our model. 
Second, we hypothesized that explicitly training our model using explanations that the user provides can help it reason in a manner similar to the user, not only improving the model's explanation performance, but also its prediction performance. To test this, we compare our model against `No Concept Training' version of our model, by removing the auxiliary explanation loss on the attention weights. This reduces the accuracy of explanations from 73.6\% to 26.4\%, and the prediction accuracy from 70.9\% to 64.4\%.

\section{Robot Demonstration}

\begin{figure}[h]
    \centering
    \vspace{-8pt}
    \includegraphics[width=0.45\textwidth]{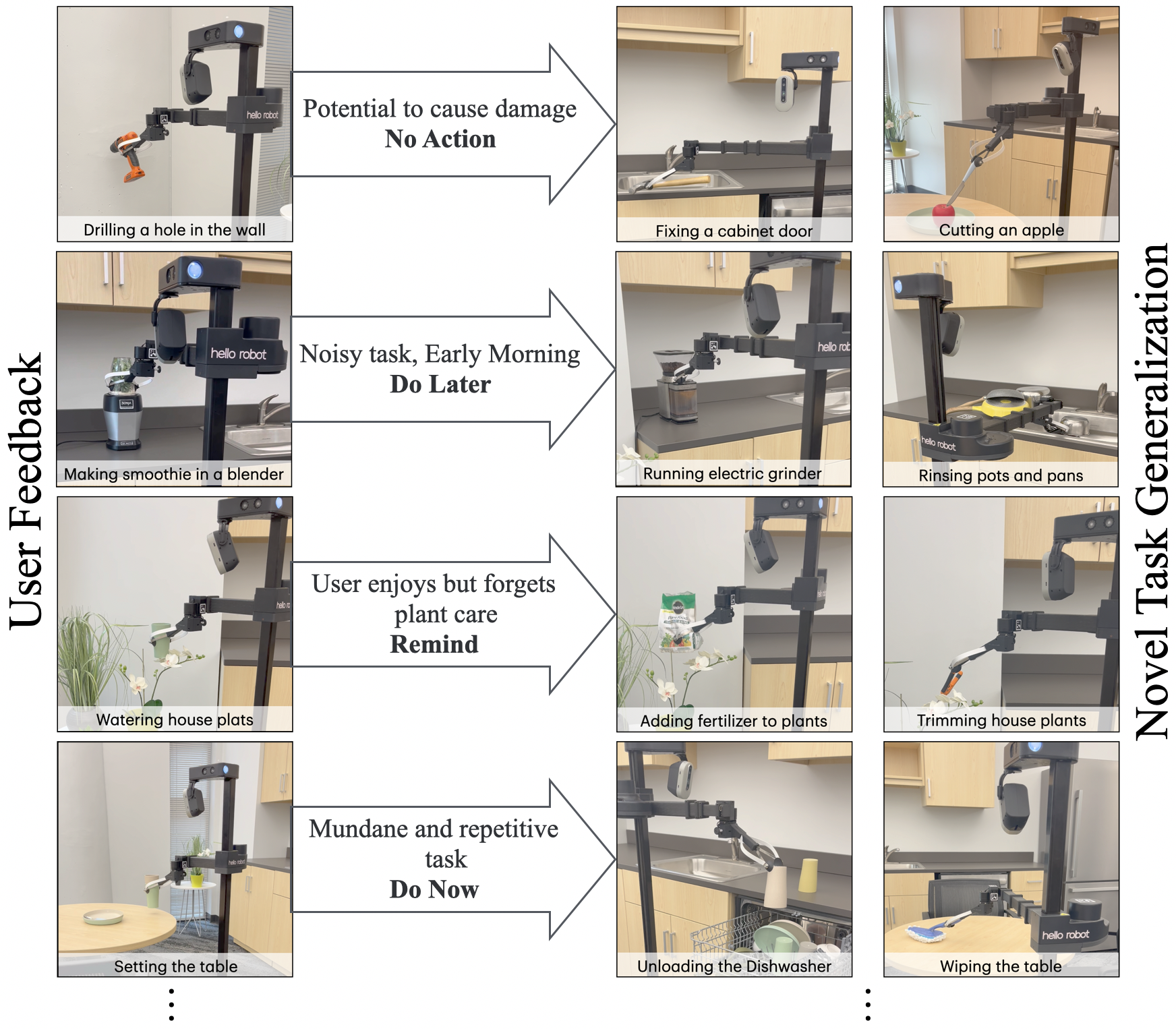}
    \setlength{\abovecaptionskip}{-2pt}
    \setlength{\belowcaptionskip}{-15pt}
    \caption{\small{Robot demo generalizing user feedback for different preferred task adaptations to novel household tasks.}}
    \label{fig:demo}
\end{figure}

Finally, we show a proof-of-concept of our system on a stretch robot\footnote{Full demo available at \url{https://youtu.be/3MiIx8GAtdw}}, showcasing personalization across a diverse array of tasks and associated concepts, as shown in the video and Figure~\ref{fig:demo}. 
With our framework, the robot can accept feedback over a few tasks to customize behavior, and generalize to a wider set of tasks across the household.

\section{Conclusion and Limitations}

In this work, we propose TAACo, which can help adapt robot behavior to user preferences by picking between four ways of assisting with a task: doing it now, doing it later, reminding the user about it, or not doing anything. Our core contribution is the use of abstract concept-based representation and training, which allows generalization to an open-set of household tasks and enables explainability. However, our work has some limitations, opening avenues for future work. First, our input task representation includes a textual action and activity description, and a set of objects and locations, which cannot capture details, such as the role of an object in the given task, which might be pertinent to some preferences. Second, for some individuals or environments, the notion of component-concept alignment might diverge from the norm. Since we do not customize the commonsense module, we cannot model such preferences. Finally, future work can explore ideas of modeling change in user preferences, and lifelong learning in context of this problem.

\vspace{-3pt}
\section*{Acknowledgment}
\vspace{-5pt}
\noindent Sponsored in part by NSF IIS 2112633 and Amazon Research.
\vspace{-14pt}

\bibliographystyle{ieeetr}
\bibliography{references}

\begin{thebibliography}{10}

\bibitem{jiang2022vima}
Y.~Jiang, A.~Gupta, Z.~Zhang, G.~Wang, Y.~Dou, Y.~Chen, L.~Fei-Fei, A.~Anandkumar, Y.~Zhu, and L.~Fan, ``Vima: General robot manipulation with multimodal prompts,'' in {\em ICML}, 2023.

\bibitem{shridhar2023perceiver}
M.~Shridhar, L.~Manuelli, and D.~Fox, ``Perceiver-actor: A multi-task transformer for robotic manipulation,'' pp.~785--799, PMLR, 2023.

\bibitem{lynch2023interactive}
C.~Lynch, A.~Wahid, J.~Tompson, T.~Ding, J.~Betker, R.~Baruch, T.~Armstrong, and P.~Florence, ``Interactive language: Talking to robots in real time,'' {\em IEEE RAL}, 2023.

\bibitem{liang2024learning}
$-$, ``Learning to learn faster from human feedback with language model predictive control,'' 2024.

\bibitem{patel2022proactive}
M.~Patel and S.~Chernova, ``Proactive robot assistance via spatio-temporal object modeling,'' vol.~205, pp.~881--891, PMLR, 2023.

\bibitem{patel2023predicting}
M.~Patel, A.~G. Prakash, and S.~Chernova, ``Predicting routine object usage for proactive robot assistance,'' vol.~229, PMLR, 2023.

\bibitem{mascaro2023hoiabot}
E.~V. Mascaro, D.~Sliwowski, and D.~Lee, ``{HOI}4{ABOT}: Human-object interaction anticipation for human intention reading collaborative ro{BOT}s,'' in {\em 7th Annual}, 2023.

\bibitem{kapelyukh2022my}
I.~Kapelyukh and E.~Johns, ``My house, my rules: Learning tidying preferences with graph neural networks,'' pp.~740--749, PMLR, 2022.

\bibitem{jain2023transformers}
V.~Jain, Y.~Lin, E.~Undersander, Y.~Bisk, and A.~Rai, ``Transformers are adaptable task planners,'' pp.~1011--1037, PMLR, 2023.

\bibitem{peng2023diagnosis}
A.~Peng, A.~Netanyahu, M.~K. Ho, T.~Shu, A.~Bobu, J.~Shah, and P.~Agrawal, ``Diagnosis, feedback, adaptation: A human-in-the-loop framework for test-time policy adaptation,'' in {\em ICML}, 2023.

\bibitem{yuan2022situ}
L.~Yuan, X.~Gao, Z.~Zheng, M.~Edmonds, Y.~N. Wu, F.~Rossano, H.~Lu, Y.~Zhu, and S.-C. Zhu, ``In situ bidirectional human-robot value alignment,'' {\em Science robotics}, vol.~7, 2022.

\bibitem{ranz2017capability}
F.~Ranz, V.~Hummel, and W.~Sihn, ``Capability-based task allocation in human-robot collaboration,'' {\em Procedia Manufacturing}, vol.~9, 2017.

\bibitem{malik2019complexity}
A.~A. Malik and A.~Bilberg, ``Complexity-based task allocation in human-robot collaborative assembly,'' {\em Industrial Robot: the international journal of robotics research and application}, vol.~46, 2019.

\bibitem{gjeldum2022collaborative}
N.~Gjeldum, A.~Aljinovic, M.~Crnjac~Zizic, and M.~Mladineo, ``Collaborative robot task allocation on an assembly line using the decision support system,'' {\em International Journal of Computer Integrated Manufacturing}, vol.~35, 2022.

\bibitem{zhao2023learning}
M.~D. Zhao, R.~Simmons, and H.~Admoni, ``Learning human contribution preferences in collaborative human-robot tasks,'' in {\em CORL}, 2023.

\bibitem{smarr2014domestic}
C.-A. Smarr, T.~L. Mitzner, J.~M. Beer, A.~Prakash, T.~L. Chen, C.~C. Kemp, and W.~A. Rogers, ``Domestic robots for older adults: attitudes, preferences, and potential,'' {\em IJSC}, vol.~6, 2014.

\bibitem{saunders2015teach}
J.~Saunders, D.~S. Syrdal, K.~L. Koay, N.~Burke, and K.~Dautenhahn, ``“teach me--show me”—end-user personalization of a smart home and companion robot,'' {\em HMS}, vol.~46, 2015.

\bibitem{leonardi2019trigger}
N.~Leonardi, M.~Manca, F.~Patern{\`o}, and C.~Santoro, ``Trigger-action programming for personalising humanoid robot behaviour,'' in {\em CHI}, 2019.

\bibitem{kubota2020jessie}
A.~Kubota, E.~I. Peterson, V.~Rajendren, H.~Kress-Gazit, and L.~D. Riek, ``Jessie: Synthesizing social robot behaviors for personalized neurorehabilitation and beyond,'' in {\em HRI}, 2020.

\bibitem{zhuang2020consequences}
S.~Zhuang and D.~Hadfield-Menell, ``Consequences of misaligned ai,'' {\em NeurIPS}, vol.~33, 2020.

\bibitem{skalse22defining}
J.~Skalse, N.~Howe, D.~Krasheninnikov, and D.~Krueger, ``Defining and characterizing reward gaming,'' in {\em NeurIPS}, vol.~35, 2022.

\bibitem{cui2023no}
Y.~Cui, S.~Karamcheti, R.~Palleti, N.~Shivakumar, P.~Liang, and D.~Sadigh, ``No, to the right: Online language corrections for robotic manipulation via shared autonomy,'' in {\em HRI}, 2023.

\bibitem{sharma2022correcting}
P.~Sharma, B.~Sundaralingam, V.~Blukis, C.~Paxton, T.~Hermans, A.~Torralba, J.~Andreas, and D.~Fox, ``Correcting robot plans with natural language feedback,'' {\em arXiv}, 2022.

\bibitem{cui2020empathic}
Y.~Cui, Q.~Zhang, B.~Knox, A.~Allievi, P.~Stone, and S.~Niekum, ``The empathic framework for task learning from implicit human feedback,'' vol.~155, PMLR, 2021.

\bibitem{ramachandruni2023consor}
K.~Ramachandruni, M.~Zuo, and S.~Chernova, ``Consor: A context-aware semantic object rearrangement framework for partially arranged scenes,'' in {\em IROS}, IEEE, 2023.

\bibitem{lindner2021learning}
D.~Lindner, R.~Shah, P.~Abbeel, and A.~Dragan, ``Learning what to do by simulating the past,'' {\em arXiv}, 2021.

\bibitem{munzer2017preference}
T.~Munzer, M.~Toussaint, and M.~Lopes, ``Preference learning on the execution of collaborative human-robot tasks,'' in {\em ICRA}, IEEE, 2017.

\bibitem{zhao2022coordination}
M.~Zhao, R.~Simmons, and H.~Admoni, ``Coordination with humans via strategy matching,'' in {\em IROS}, IEEE, 2022.

\bibitem{nemlekar2023transfer}
H.~Nemlekar, N.~Dhanaraj, A.~Guan, S.~K. Gupta, and S.~Nikolaidis, ``Transfer learning of human preferences for proactive robot assistance in assembly tasks,'' in {\em ACM/IEEE HRI}, 2023.

\bibitem{hadfield2016cooperative}
D.~Hadfield-Menell, S.~J. Russell, P.~Abbeel, and A.~Dragan, ``Cooperative inverse reinforcement learning,'' {\em NeurIPS}, vol.~29, 2016.

\bibitem{du2020ave}
Y.~Du, S.~Tiomkin, E.~Kiciman, D.~Polani, P.~Abbeel, and A.~Dragan, ``Ave: Assistance via empowerment,'' {\em Advances in Neural Information Processing Systems}, vol.~33, pp.~4560--4571, 2020.

\bibitem{quartey2024verifiably}
B.~Quartey, E.~Rosen, S.~Tellex, and G.~Konidaris, ``Verifiably following complex robot instructions with foundation models,'' {\em arXiv}, 2024.

\bibitem{song2023llm}
C.~H. Song, J.~Wu, C.~Washington, B.~M. Sadler, W.-L. Chao, and Y.~Su, ``Llm-planner: Few-shot grounded planning for embodied agents with large language models,'' in {\em IEEE/CVF CVPR}, pp.~2998--3009, 2023.

\bibitem{chen2023open}
B.~Chen, F.~Xia, B.~Ichter, K.~Rao, K.~Gopalakrishnan, M.~S. Ryoo, A.~Stone, and D.~Kappler, ``Open-vocabulary queryable scene representations for real world planning,'' in {\em ICRA}, IEEE, 2023.

\bibitem{huang2023visual}
C.~Huang, O.~Mees, A.~Zeng, and W.~Burgard, ``Visual language maps for robot navigation,'' in {\em ICRA}, IEEE, 2023.

\bibitem{sobrin2024explaining}
D.~Sobr{\'\i}n-Hidalgo, M.~A. Gonz{\'a}lez-Santamarta, {\'A}.~M. Guerrero-Higueras, F.~J. Rodr{\'\i}guez-Lera, and V.~Matell{\'a}n-Olivera, ``Explaining autonomy: Enhancing human-robot interaction through explanation generation with large language models,'' {\em arXiv}, 2024.

\bibitem{wu2023tidybot}
J.~Wu, R.~Antonova, A.~Kan, M.~Lepert, A.~Zeng, S.~Song, J.~Bohg, S.~Rusinkiewicz, and T.~Funkhouser, ``Tidybot: Personalized robot assistance with large language models,'' {\em arXiv}, 2023.

\bibitem{zhang2023large}
B.~Zhang and H.~Soh, ``Large language models as zero-shot human models for human-robot interaction,'' in {\em IROS}, IEEE, 2023.

\bibitem{brohan2022rt1}
A.~Brohan, N.~Brown, J.~Carbajal, Y.~Chebotar, J.~Dabis, C.~Finn, K.~Gopalakrishnan, K.~Hausman, A.~Herzog, J.~Hsu, {\em et~al.}, ``Rt-1: Robotics transformer for real-world control at scale,'' {\em arXiv}, 2022.

\bibitem{booth2016mixed}
K.~E.~C. Booth, T.~T. Tran, G.~Nejat, and J.~C. Beck, ``Mixed-integer and constraint programming techniques for mobile robot task planning,'' {\em IEEE RAL}, vol.~1, 2016.

\bibitem{dantam2016incremental}
N.~T. Dantam, Z.~K. Kingston, S.~Chaudhuri, and L.~E. Kavraki, ``Incremental task and motion planning: A constraint-based approach.,'' in {\em RSS}, vol.~12, 2016.

\bibitem{das2023state2explanation}
D.~Das, S.~Chernova, and B.~Kim, ``State2explanation: Concept-based explanations to benefit agent learning and user understanding,'' 2023.

\bibitem{zabounidis2023concept}
R.~Zabounidis, J.~Campbell, S.~Stepputtis, D.~Hughes, and K.~P. Sycara, ``Concept learning for interpretable multi-agent reinforcement learning,'' pp.~1828--1837, PMLR, 2023.

\bibitem{achiam2023gpt}
J.~Achiam, S.~Adler, S.~Agarwal, L.~Ahmad, I.~Akkaya, F.~L. Aleman, D.~Almeida, J.~Altenschmidt, S.~Altman, S.~Anadkat, {\em et~al.}, ``Gpt-4 technical report,'' {\em arXiv}, 2023.

\end{thebibliography}

\end{document}